%% file: main.tex
\documentclass[sigconf]{acmart}

\microtypesetup{expansion=false}
\usepackage{booktabs}
\usepackage{multirow}
\usepackage{diagbox}
\usepackage{graphicx}
\usepackage{rotating}
\usepackage{float}
\usepackage{amsmath}
\usepackage{threeparttable}
\usepackage{makecell}
\usepackage{pifont}

\usepackage{amssymb}
\usepackage[table]{xcolor}

\newcommand{\modelns}{\textsc{UnmixDrift}}

\definecolor{UMBlue}{HTML}{3F6682}
\definecolor{UMBlueLight}{HTML}{EEF4F8}
\definecolor{UMGroup}{HTML}{F5F7F9}

\AtBeginDocument{%
}

\copyrightyear{2027}
\acmYear{2027}

\setcopyright{cc}
\setcctype{by}

\acmConference[WSDM 2027]
{Proceedings of the Twentieth ACM International Conference on Web Search and Data Mining}
{February 15--19, 2027}
{Hong Kong, Hong Kong.}

\acmBooktitle{Proceedings of the Twentieth ACM International Conference on Web Search and Data Mining (WSDM 2027), February 15--19, 2027, Hong Kong, Hong Kong}

\acmISBN{979-8-4007-3117-4/2027/02}
\acmDOI{10.1145/3862322.3863638}

\begin{document}

\title{Unmix Drift: Component-Wise Drift Representation for Online Time Series Forecasting under Delayed Feedback}

%
%
%


\author{Jierui Li}
\orcid{0009-0007-3540-2279}
\authornote{These authors contributed equally to this work.}
\authornote{This work was conducted while Jierui Li was a research intern at the State Key Laboratory of Cognitive Intelligence, University of Science and Technology of China.}
\affiliation{%
  \institution{University of Science and Technology of China}
  \city{Hefei}
  \state{Anhui}
  \country{China}
}
\email{13031692900@163.com}

\author{Zhuodong Liu}
\orcid{0009-0009-9535-2941}
\authornotemark[2]
\affiliation{%
  \institution{The Hong Kong University of Science and Technology (Guangzhou)}
  \city{Guangzhou}
  \state{\mbox{Guangdong}}
  \country{China}
}
\email{zliu209@connect.hkust-gz.edu.cn}

\author{Weidong Tang}
\orcid{0009-0009-7138-4543}
\affiliation{%
  \institution{Xidian University}
  \city{Xi'an}
  \state{Shaanxi}
  \country{China}
}
\email{wdtang0705@gmail.com}

\author{Jixian Liu}
\orcid{0009-0006-3295-6404}
\affiliation{%
  \institution{Xidian University}
  \city{Xi'an}
  \state{Shaanxi}
  \country{China}
}
\email{jixianliu@stu.xidian.edu.cn}

\author{Shuning Zhang}
\orcid{0000-0002-4145-117X}
\affiliation{%
  \institution{Tsinghua University}
  \city{Beijing}
  \country{China}
}
\email{zsn23@mails.tsinghua.edu.cn}

\author{Zhi Zheng}
\orcid{0000-0001-7758-8904}
\correspondingauthor
\affiliation{%
  \institution{University of Science and Technology of China}
  \city{Hefei}
  \state{Anhui}
  \country{China}
}
\email{zhengzhi97@ustc.edu.cn}

\author{Lingxing Kong}
\orcid{0009-0009-5087-9125}
\affiliation{%
  \institution{Tsinghua University}
  \city{Beijing}
  \country{China}
}
\email{klxstar@126.com}

\renewcommand{\shortauthors}{Li et al.}


\begin{abstract}
Online time series forecasting (OTSF) needs to adapt to concept drift under delayed feedback, where complete labels for an $H$-step forecast become available only after $H$ time steps. Existing proactive methods such as PROCEED reduce this adaptation lag by comparing global representations of the current input and the latest historical reference. However, early aggregation may obscure component-level current--historical discrepancies and limit the resulting adaptation signal.
To this end, we propose \textbf{Unmix Drift}, a proactive adaptation framework with component-wise drift construction. It follows a \emph{decompose--compare--aggregate} strategy that preserves component-level discrepancies until final aggregation, combining normalized component-wise residuals with salience derived from pre-normalization response changes to form a compact drift representation for lightweight parameter modulation. Because component decomposition depends on the learned representation while parameter generation consumes the resulting drift signal, jointly training all modules from the outset exposes downstream modules to continuously changing intermediate representations. We therefore adopt a stage-wise warm-up strategy that sequentially refines the projection, decomposition, and generation modules in dependency order, providing more stable inputs to downstream stages.
Experiments across three forecasting backbones, five real-world datasets, and three forecasting horizons show that Unmix Drift achieves lower MSE than PROCEED in 44 of 45 configurations and ranks first or second in all configurations on both MSE and MAE, while maintaining comparable per-step latency and GPU memory. The code is available at \url{https://github.com/unmixd/unmix-drift}.
\end{abstract}



\maketitle

\input{sec/1.introduction}
\input{sec/2.related_works}
\input{sec/3.preliminary}
\input{sec/4.technique}

\input{tab/tables_synced}

\input{sec/5.experiment}

\input{sec/6.conclusion}

\begin{acks}
This work was supported in part by grants from the
\grantsponsor{nsfc}{National Natural Science Foundation of China}{https://doi.org/10.13039/501100001809}
(Grant No.~\grantnum{nsfc}{62502480}),
the \grantsponsor{postdoc}{Postdoctoral Fellowship Program}{https://www.chinapostdoctor.org.cn/}
(Grant No.~\grantnum{postdoc}{BX20250387}),
the \grantsponsor{cpsf}{China Postdoctoral Science Foundation}{https://www.chinapostdoctor.org.cn/}
(Grant No.~\grantnum{cpsf}{2025M781529}), and
the \grantsponsor{frfcu}{Fundamental Research Funds for the Central Universities}{https://finance.ustc.edu.cn/}
(Grant No.~\grantnum{frfcu}{WK2150250042}).
\end{acks}

\section*{Ethical Considerations}

This work studies online time-series forecasting using established
benchmark datasets and does not involve human subjects or the
collection of personally identifiable information. As forecasting
models may be used in real-world decision-making systems, inaccurate
predictions could lead to undesirable downstream decisions.
Accordingly, models developed under the studied setting should be
carefully validated for the target domain and used with appropriate
human or system-level oversight in high-stakes applications.



\bibliographystyle{ACM-Reference-Format}
\bibliography{custom}

\end{document}

%% file: sec/1.introduction.tex
\section{Introduction}

\begin{figure*}
\centering
\includegraphics[width=\linewidth]{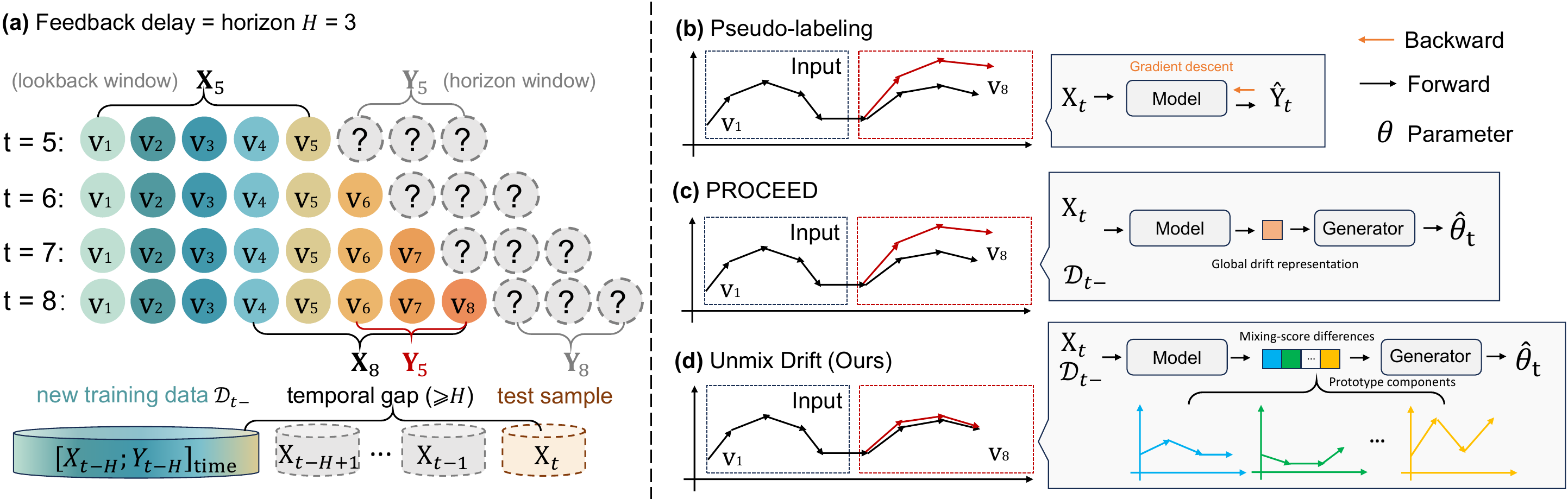}
\Description{
Four-panel comparison of online adaptation strategies.
(a) H-step feedback delay.
(b) Pseudo-labeling with surrogate supervision.
(c) PROCEED using global drift representations.
(d) Unmix Drift using component-wise differences.
}
\caption{
Comparison of online adaptation strategies under delayed feedback.
(a) $H$-step feedback delay.
(b) Pseudo-labeling: adaptation using surrogate supervision before complete labels arrive.
(c) PROCEED forms global concept representations before computing current--historical drift.
(d) Unmix Drift preserves component-wise variation through component-wise mixing-score differences before final aggregation and parameter generation.
}
\label{fig:intro}
\end{figure*}

Online time series forecasting (OTSF) continuously updates forecasting models as new observations arrive, allowing them to adapt to non-stationary data streams \cite{pham2023learning,wen2023onenet,zhao2025proactive}. Time-series forecasting supports applications in energy, weather, traffic, finance, and industrial sensing \cite{zhou2021informer,wu2020connecting,lyu2025timing,zhang2025multi,zhang2025camef,tu2025astnet}. Distributional changes in evolving streams are commonly studied under the notion of \textit{concept drift} \cite{gama2014survey,lu2019learning}. Multi-step online forecasting, however, is inherently subject to \textit{feedback delay}. For a forecasting horizon of $H$, the complete target associated with a prediction at time $t$ becomes available only after the following $H$ time steps \cite{zhao2025proactive}. The latest complete supervision therefore lags behind the current query, while a strict online protocol requires each prediction to be made before its target becomes available to avoid future-information leakage \cite{lau2025fast,liang2024act,zhao2025proactive}. Figure~\ref{fig:intro}(a) illustrates how feedback-only adaptation lags behind the current data state.

Existing methods reduce this lag through different forms of early adaptation. Some exploit incomplete, partial, or surrogate supervision before full feedback becomes available \cite{lau2025fast,liang2024act}, as illustrated in Figure~\ref{fig:intro}(b). A complementary direction is \textit{proactive adaptation}, which uses the current unlabeled input to adjust the model before its target arrives. PROCEED \cite{zhao2025proactive}, for example, aggregates latent features of the current input and a historical reference into global concept representations, computes their difference, and uses the resulting drift representation to generate parameter adjustments, as illustrated in Figure~\ref{fig:intro}(c). This global representation is effective, but component-level current--historical discrepancies are not explicitly retained before aggregation. This raises a more specific \textit{representation question}: \textit{how should the current--historical difference be represented before it is used for proactive adaptation?} 

To address this question, we propose \textbf{Unmix Drift}, a component-wise proactive adaptation framework for online forecasting under delayed feedback. As illustrated in Figure~\ref{fig:intro}(d), Unmix Drift follows a \textit{decompose--compare--aggregate} process. The current query and the latest fully resolved labeled context are mapped to a shared set of learned components. Normalized component-wise residuals retain signed current--historical discrepancies, while relative changes in the pre-normalization responses provide component salience. Salience-weighted aggregation then converts these component-level discrepancies into a compact drift representation that drives lightweight parameter modulation before the current label becomes available.

The framework also introduces a natural dependency across its trainable modules. Component decomposition operates on the learned representation space, while parameter generation consumes the resulting drift representation. Joint training from the outset therefore exposes downstream modules to intermediate representations that continue to change as upstream modules are optimized. To address this, we adopt a \textit{stage-wise warm-up} strategy that sequentially refines the projection, decomposition, and generation modules according to their dependencies, yielding a more stable drift representation for downstream parameter generation.

We evaluate Unmix Drift on three forecasting backbones, five real-world datasets, and three forecasting horizons, resulting in 45 backbone--dataset--horizon configurations under a strict leakage-free delayed-feedback protocol. Unmix Drift achieves lower MSE than PROCEED in 44 of 45 configurations and ranks first or second in every configuration on both MSE and MAE. Ablation studies examine the drift construction and training strategy, representation analyses provide diagnostics of the learned adaptation signal, and efficiency measurements show per-step latency and GPU memory comparable to PROCEED.

Our main contributions are summarized as follows:
\begin{itemize}

\item We introduce a component-wise drift representation for proactive adaptation under delayed feedback. Unmix Drift follows a \textit{decompose--compare--aggregate} process that preserves current--historical discrepancies until final aggregation and forms a compact signal for parameter modulation.

\item We combine salience-weighted component aggregation with a stage-wise warm-up strategy that sequentially refines the projection, decomposition, and generation modules according to their dependencies, yielding a more stable drift representation for downstream parameter generation.

\item We conduct a leakage-free evaluation across 45 backbone--dataset--horizon settings. Unmix Drift outperforms PROCEED in MSE in 44 of 45 settings and ranks first or second on both MSE and MAE, with comparable per-step latency and GPU memory.

\end{itemize}

%% file: sec/2.related_works.tex
\section{Related Work}

\subsection{Online Time-Series Forecasting under Concept Drift}

Online time-series forecasting (OTSF) operates on continuously arriving non-stationary streams and requires models to adapt to evolving distributions during deployment. Beyond TCN~\cite{bai2018empirical}, PatchTST~\cite{nie2022time}, and iTransformer~\cite{liu2024itransformerinvertedtransformerseffective}, representative architectures include Autoformer~\cite{autoformer}, TimesNet~\cite{timesnet}, DUET~\cite{qiu2025duet}, and ST-MTM~\cite{seo2025st}, exploring sequence decomposition and auto-correlation, multi-period modeling, dual clustering across temporal and channel dimensions, and masked time-series modeling with seasonal--trend decomposition, respectively.

Distribution shift has also been studied explicitly in time-series modeling. RevIN~\cite{revin} mitigates changes in time-varying statistics through reversible instance normalization, while Non-stationary Transformers~\cite{nonstationary_transformer} retain non-stationary information alongside stationarization. AdaRNN~\cite{du2021adarnn} addresses temporal covariate shift, while non-stationary time-aware kernelized attention~\cite{ma2022non} models temporal dependencies in event prediction. ADWIN~\cite{adwin} detects distribution changes in data streams through adaptive windowing, while You et al.~\cite{you2021learning} explicitly model future concept changes in time-series prediction.

Within online forecasting, FSNet~\cite{pham2023learning} introduces fast--slow learning to balance rapid responses with retention of historical patterns. OneNet~\cite{wen2023onenet} dynamically ensembles forecasting models, while SOLID~\cite{chen2024calibration} retrieves contextually similar historical samples for lightweight prediction-layer calibration. These studies address forecasting non-stationarity through architecture design, distribution adaptation, or online updating, but their original settings do not explicitly center on the strict $H$-step availability constraint of complete supervision in multi-step online forecasting.

\subsection{Leakage-Free Online Forecasting under Delayed Feedback}

In multi-step online forecasting, a forecasting horizon of $H$ naturally delays the availability of the complete ground-truth target. Under a strict leakage-free setting, model updates can therefore only use information genuinely available at the current time. Recent studies have explored leakage-free updating under incomplete or delayed supervision. DSOF~\cite{lau2025fast} adopts a fast--slow dual-stream mechanism: the fast stream uses temporal difference learning for recent adaptation, while the slow stream updates with complete data through experience replay. Act-Now~\cite{liang2024act} uses fast and slow buffers for immediate updates with consistent pseudo and partial labels and parallel updates with earlier complete labels.

PROCEED~\cite{zhao2025proactive} further performs proactive adaptation by exploiting the difference between the current unlabeled query and the latest fully supervised historical reference. It first encodes and aggregates the two inputs into global concept representations, computes their vector difference as a drift representation, and uses the resulting representation to generate model parameter adjustments. Building on this direction, we study how to retain component-wise current--historical discrepancies before final aggregation so that a more structured drift signal can be provided for proactive adaptation.

%% file: sec/3.preliminary.tex
\section{Problem Formulation}
\label{sec:problem}

\begin{figure*}[t]
    \centering
    \includegraphics[width=0.92\textwidth]{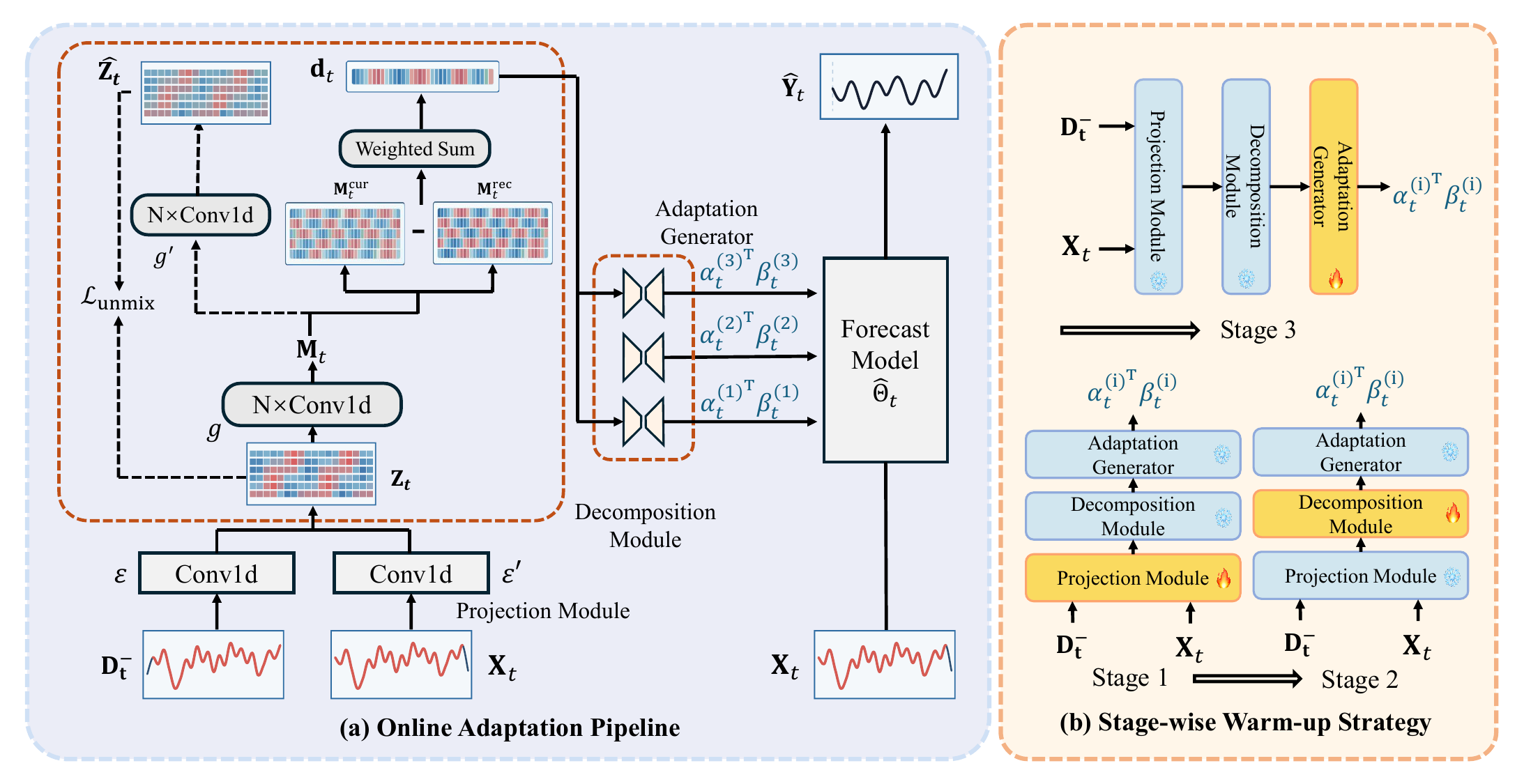}
    \Description{
        Two-panel overview of Unmix Drift.
        (a) Component-wise drift construction and
        parameter adaptation under delayed feedback.
        (b) Three-stage warm-up of the projection,
        decomposition, and adaptation-generator modules.
    }
    \caption{
        Overview of \textbf{Unmix Drift}.
        (a) Component-wise drift representation and parameter adaptation
        under delayed feedback; the pre-normalization salience path is
        omitted for clarity.
        (b) Stage-wise warm-up of the projection, decomposition, and
        adaptation-generator modules.
    }
    \label{fig:pipeline}
\end{figure*}

\textbf{Online forecasting.}
Let $\{\mathbf{v}_t\}_{t\ge 1}$ denote a continuously arriving stream of
$N$-variate observations, where $\mathbf{v}_t\in\mathbb{R}^{N}$.
At each time step $t$, the model receives a lookback window
$\mathbf{X}_t=[\mathbf{v}_{t-L+1},\dots,\mathbf{v}_t]
\in\mathbb{R}^{L\times N}$
and predicts the future horizon
$\mathbf{Y}_t=[\mathbf{v}_{t+1},\dots,\mathbf{v}_{t+H}]
\in\mathbb{R}^{H\times N}$
using a forecasting model $f(\cdot;\Theta)$ parameterized by $\Theta$.

In multi-step online forecasting, $\mathbf{Y}_t$ becomes fully observable
only at time $t+H$~\cite{yasui2022learning,csaba2024label}. Hence, the
latest complete supervision available at decision time $t$ is
\begin{equation}
\mathcal{R}_t^-=
(\mathbf{X}_{t-H},\mathbf{Y}_{t-H}).
\label{eq:latest_feedback}
\end{equation}

Let $\mathcal{U}$ denote the causal delayed-feedback update of the
underlying backbone. Before proactive modulation, the backbone state is
updated as
\begin{equation}
\Theta_t^-=
\mathcal{U}(\Theta_{t-1}^-,\mathcal{R}_t^-),
\label{eq:causal_base_update}
\end{equation}
where no observation after time $t$ is used. Although $\Theta_t^-$ is
indexed by the current decision time $t$, its latest fully supervised
forecast episode originates at $t-H$. Without proactive adaptation, the
current prediction is
\begin{equation}
\widetilde{\mathbf{Y}}_t=
f(\mathbf{X}_t;\Theta_t^-).
\label{eq:delayed_backbone_pred}
\end{equation}

\textbf{Delayed feedback and resolved context.}
Since $\mathbf{Y}_{t-H}$ is fully observed at time $t$, we concatenate it
with its associated input along the temporal dimension to form the latest
fully resolved labeled context,
\begin{equation}
\mathbf{D}_t^-=
[\mathbf{X}_{t-H};\mathbf{Y}_{t-H}]_{\mathrm{time}}
\in\mathbb{R}^{(L+H)\times N}.
\label{eq:delayed_reference}
\end{equation}

By construction,
$\mathbf{D}_t^-=
[\mathbf{v}_{t-H-L+1},\dots,\mathbf{v}_t]$,
so $\mathbf{X}_t$ is the most recent $L$-step suffix of
$\mathbf{D}_t^-$. The two therefore share observations but serve
different causal roles: $\mathbf{D}_t^-$ is formed from the latest
forecast episode whose target has become fully resolved, whereas
$\mathbf{X}_t$ is the current query whose future target remains
unavailable. Their distinction is causal rather than based on temporally
disjoint observations.

In a non-stationary stream, the distribution associated with the latest
available supervision,
$P(\mathbf{X}_{t-H},\mathbf{Y}_{t-H})$,
may differ from that of the current forecasting episode,
$P(\mathbf{X}_t,\mathbf{Y}_t)$. Adaptation based only on delayed
supervision can therefore lag behind the current data state. Unmix Drift
constructs a representation of this resolved-context--query discrepancy
and applies temporary, query-conditioned modulation on top of
$\Theta_t^-$ to obtain $\widehat{\Theta}_t$ for the current prediction.

%% file: sec/4.technique.tex
\section{The Proposed Framework}

Unmix Drift comprises three parts:
(i) component-wise drift representation,
(ii) drift-driven adapter-parameter generation and injection,
and (iii) stage-wise warm-up on the validation stream.

\subsection{Component-Wise Drift Representation}
\label{subsec:prototype_unmixing}

Before the current target becomes available, the model needs to represent how
the current query differs from the latest fully resolved labeled context.
We construct a component-wise drift representation from $\mathbf{X}_t$ and
$\mathbf{D}_t^-$ that preserves component-level current--historical
discrepancies before final aggregation. As illustrated in
Figure~\ref{fig:pipeline}(a), the process consists of latent-space projection,
component-response construction, current--historical comparison, and
salience-weighted aggregation.

The current query and resolved context have different temporal extents and
are first mapped into a comparable representation space:
\begin{equation}
\mathbf{H}_{\mathrm{cur}}
=
\varepsilon'(\mathbf{X}_t),
\qquad
\mathbf{H}_{\mathrm{rec}}
=
\varepsilon(\mathbf{D}_t^-),
\label{eq:projection_representation}
\end{equation}
where
$\mathbf{H}_{\mathrm{cur}},
\mathbf{H}_{\mathrm{rec}}
\in\mathbb{R}^{1\times N\times D}$,
and $D$ is the latent feature dimension.
We concatenate them along the current--reference pair dimension:
\begin{equation}
\mathbf{Z}_t
=
[\mathbf{H}_{\mathrm{cur}};
\mathbf{H}_{\mathrm{rec}}]_{\mathrm{pair}}
\in
\mathbb{R}^{2\times N\times D}.
\label{eq:joint_latent}
\end{equation}

A score encoder $g(\cdot)$ characterizes $\mathbf{Z}_t$ over $K$ learnable
components. These components define a shared decomposition basis without
assuming semantic correspondence to physical sources of concept drift.
The decomposition module consists of the score encoder $g(\cdot)$,
a reconstruction decoder $g'(\cdot)$, and a nonlinear residual branch
$h(\cdot)$, implemented with lightweight $1\times1$ convolutions and
nonlinear activations. The score encoder produces non-negative responses:
\begin{equation}
\widetilde{\mathbf{M}}_t
=
|g(\mathbf{Z}_t)|.
\label{eq:raw_mixing_score}
\end{equation}

We normalize the responses along the component dimension:
\begin{equation}
\mathbf{M}_{t,i,k,d}
=
\frac{
\widetilde{\mathbf{M}}_{t,i,k,d}
}{
\sum_{k'=1}^{K}
\widetilde{\mathbf{M}}_{t,i,k',d}
+\eta
},
\label{eq:normalized_mixing_score}
\end{equation}
where $i\in\{\mathrm{cur},\mathrm{rec}\}$,
$k\in\{1,\ldots,K\}$, $d\in\{1,\ldots,D\}$,
and $\eta$ is a small stability constant.
The resulting
$\mathbf{M}_t\in\mathbb{R}^{2\times K\times D}$
provides comparable mixing scores over the same learned components.

To retain the directional structure of the latent representations, we
introduce an auxiliary reconstruction constraint:
\begin{equation}
\mathbf{U}_t
=
g'(\mathbf{M}_t),
\qquad
\widehat{\mathbf{Z}}_t
=
h(\mathbf{U}_t)
+
\mathbf{U}^{(1)}_t
+
\mathbf{U}^{(2)}_t,
\label{eq:unmix_reconstruction}
\end{equation}
where $\mathbf{U}^{(1)}_t$ and $\mathbf{U}^{(2)}_t$ are obtained by
splitting $\mathbf{U}_t$ along the channel dimension.

For $i\in\{\mathrm{cur},\mathrm{rec}\}$ and latent dimension $d$, the cosine
similarity between the original and reconstructed representations is
\begin{equation}
c_{t,i,d}
=
\frac{
\mathbf{Z}_{t,i,:,d}^{\top}
\widehat{\mathbf{Z}}_{t,i,:,d}
}{
\|\mathbf{Z}_{t,i,:,d}\|_2
\|\widehat{\mathbf{Z}}_{t,i,:,d}\|_2+\delta
},
\label{eq:unmix_cosine}
\end{equation}
and the reconstruction objective is the mean angular error
\begin{equation}
\mathcal{L}_{\mathrm{unmix}}
=
\frac{1}{2D}
\sum_{i\in\{\mathrm{cur},\mathrm{rec}\}}
\sum_{d=1}^{D}
\arccos\left(
\operatorname{clip}(c_{t,i,d},-1,1)
\right),
\label{eq:unmix_loss}
\end{equation}
where $\delta$ is a stability constant and $\operatorname{clip}(\cdot)$
guards $\arccos(\cdot)$. This objective encourages the mixing scores to
preserve directional information for subsequent current--historical
comparison.

We separate the current and resolved-context parts,
$\mathbf{M}_t=
[\mathbf{M}^{\mathrm{cur}}_t;
\mathbf{M}^{\mathrm{rec}}_t]_{\mathrm{pair}}$,
and define the component-wise residual
\begin{equation}
\Delta\mathbf{M}_t
=
\mathbf{M}^{\mathrm{cur}}_t
-
\mathbf{M}^{\mathrm{rec}}_t
\in
\mathbb{R}^{K\times D}.
\label{eq:mixing_residual}
\end{equation}
Thus, $\Delta\mathbf{M}_t$ retains component-level current--historical
variation before final aggregation.

We next compress the residual into a drift vector. The key design is to
separate \emph{change direction} from \emph{component salience}.
The normalized residual $\Delta\mathbf{M}_t$ captures signed redistribution,
while salience is estimated from pre-normalization responses to avoid the
cross-component coupling introduced by normalization. Splitting
$\widetilde{\mathbf M}_t$ into current and reference parts, we define
\begin{equation}
q_{t,k,d}
=
\frac{
\left|
\widetilde{\mathbf M}^{\mathrm{cur}}_{t,k,d}
-
\widetilde{\mathbf M}^{\mathrm{rec}}_{t,k,d}
\right|
}{
\widetilde{\mathbf M}^{\mathrm{cur}}_{t,k,d}
+
\widetilde{\mathbf M}^{\mathrm{rec}}_{t,k,d}
+\eta
}.
\label{eq:relative_raw_change}
\end{equation}
The salience of component $k$ is
\begin{equation}
\phi_{t,k}
=
\|\mathbf q_{t,k,:}\|_2,
\label{eq:prototype_drift_strength}
\end{equation}
and its weight is obtained by a temperature-controlled softmax:
\begin{equation}
w_{t,k}
=
\frac{
\exp(\phi_{t,k}/\tau)
}{
\sum_{k'=1}^{K}
\exp(\phi_{t,k'}/\tau)
}.
\label{eq:prototype_drift_weight}
\end{equation}
Here, $\tau>0$ controls the concentration of component salience.
The relative form in Eq.~\eqref{eq:relative_raw_change} reduces sensitivity
to differences in absolute response scale.

The final drift vector is obtained by salience-weighted aggregation:
\begin{equation}
\mathbf{d}_t
=
\sum_{k=1}^{K}
w_{t,k}
\Delta\mathbf{M}_{t,k,:}
\in
\mathbb{R}^{D}.
\label{eq:drift_vector}
\end{equation}
This aggregation emphasizes components with stronger relative changes while
retaining information from all components. Direction and salience are derived
from different quantities. For $K{=}2$, component-wise normalization makes
the signed residuals approximately antisymmetric,
$\Delta\mathbf M_{t,1,:}\approx-\Delta\mathbf M_{t,2,:}$, so uniform
aggregation cancels their direction. Using the two-class softmax identity,
Eq.~\eqref{eq:drift_vector} becomes
$\mathbf d_t\approx\gamma_t\,\Delta\mathbf M_{t,1,:}$, where
\[
\gamma_t=
\tanh\left(
\frac{\phi_{t,1}-\phi_{t,2}}{2\tau}
\right).
\]
Thus, the $K{=}2$ signal can be viewed as a signed redistribution direction
modulated by a salience-contrast gate. Similar salience suppresses the
resulting vector, whereas asymmetric salience retains the more strongly
weighted direction. The resulting $\mathbf{d}_t$ is passed to the adaptation
generator.

\subsection{Drift-driven Adapter Parameter Generation}
\label{sec:drift_to_adapter}

We use an adaptation generator to map the drift vector $\mathbf{d}_t$ into
layer-wise parameter modulation terms. As shown in
Figure~\ref{fig:pipeline}(a), the generator produces low-rank modulation
parameters for each adapted layer and applies them to the causally updated
backbone parameters.

Given $\mathbf{d}_t$, the generator produces
\begin{equation}
\Delta\Theta_t
=
g_{\mathrm{gen}}(\mathbf{d}_t)
=
\left\{
\boldsymbol{\alpha}^{(\ell)}_{t},
\boldsymbol{\beta}^{(\ell)}_{t}
\right\}_{\ell\in\mathcal{S}},
\label{eq:adapter_params}
\end{equation}
where $\mathcal{S}$ denotes the set of adapted submodules.
For layer $\ell$, its causal pre-modulation parameters are
$\Theta_t^{-,(\ell)}$. The generated coefficients form a low-rank rescaling
matrix:
\begin{equation}
\mathbf{A}_{t}^{(\ell)}
=
\boldsymbol{\alpha}^{(\ell)\top}_{t}
\boldsymbol{\beta}^{(\ell)}_{t}.
\label{eq:low_rank_adapter}
\end{equation}
For
$\Theta_t^{-,(\ell)}\in\mathbb R^{m_\ell\times n_\ell}$,
the generator outputs
$\boldsymbol\alpha_t^{(\ell)}\in\mathbb R^{r_\ell\times m_\ell}$ and
$\boldsymbol\beta_t^{(\ell)}\in\mathbb R^{r_\ell\times n_\ell}$,
where $r_\ell\ll\min(m_\ell,n_\ell)$.
Convolutional kernels are reshaped into matrices for modulation and restored
afterward, so Eq.~\eqref{eq:low_rank_adapter} matches the shape of the
parameter it modulates. Parameter generation and low-rank parameterization
are widely used for compact context-dependent adaptation
~\cite{ha2017hypernetworks,hu2022lora,zhang2023adalora}.

The adapted parameters are
\begin{equation}
\widehat{\Theta}_{t}^{(\ell)}
=
\mathbf{A}_{t}^{(\ell)}
\odot
\Theta_t^{-,(\ell)}
=
\left(
\boldsymbol{\alpha}^{(\ell)\top}_{t}
\boldsymbol{\beta}^{(\ell)}_{t}
\right)
\odot
\Theta_t^{-,(\ell)},
\label{eq:param_adapt}
\end{equation}
where $\odot$ denotes element-wise multiplication.
This low-rank modulation adjusts selected layer responses without directly
performing large-scale backbone updates.

The adapted parameters are then used for the current prediction:
\begin{equation}
\widehat{\mathbf{Y}}_{t}
=
f(\mathbf{X}_{t};\widehat{\Theta}_{t}).
\label{eq:adapter_inject}
\end{equation}
Compared with the direct backbone prediction in
Eq.~\eqref{eq:delayed_backbone_pred}, Unmix Drift performs
query-conditioned parameter modulation before the current label becomes
available.

\subsection{Stage-wise Warm-up on the Validation Stream}
\label{sec:stage_warmup}

The projection, decomposition, and adaptation-generator modules form a
dependency chain: decomposition operates on the learned representation,
while parameter generation consumes the resulting drift representation.
Joint optimization from the outset therefore exposes downstream modules to
changing upstream representations. We address this with stage-wise warm-up
on the validation stream. The backbone and adaptation modules are first
initialized on the training split, so frozen modules retain trained rather
than random parameters. The validation stream is then processed
chronologically under the same delayed-feedback constraint as the test
stream, without using any test observation. As shown in
Figure~\ref{fig:pipeline}(b), the projection, decomposition, and generation
modules are refined sequentially.

Whenever a validation target becomes causally available, the forecasting
objective is
\begin{equation}
\mathcal L_{\mathrm{forecast}}
=
\frac{1}{HN}
\left\|
\widehat{\mathbf Y}-\mathbf Y
\right\|_F^2.
\label{eq:forecast_loss}
\end{equation}
Stages 1 and 3 use the same forecasting loss and differ only in which module
receives gradients.

\textbf{Stage 1.}
We train the projection module while keeping the backbone, initialized
decomposition module, and adaptation generator fixed. This first learns the
representation space under the forecasting objective. The objective is
$\min_{\theta_{\mathrm{proj}}}\mathcal{L}_{\mathrm{forecast}}$, where
$\theta_{\mathrm{proj}}$ denotes the projection-encoder parameters.

\textbf{Stage 2.}
We fix the projection module and train only the decomposition module using
the angular reconstruction loss, encouraging the mixing scores to retain the
directional structure of the projected representations. The objective is
$\min_{\theta_{\mathrm{unmix}}}\mathcal{L}_{\mathrm{unmix}}$, where
$\theta_{\mathrm{unmix}}$ denotes the decomposition parameters.

\textbf{Stage 3.}
We fix the projection and decomposition modules and train only the adaptation
generator to map the drift representation into layer-wise modulation terms.
The objective is
$\min_{\theta_{\mathrm{gen}}}\mathcal{L}_{\mathrm{forecast}}$, where
$\theta_{\mathrm{gen}}$ denotes the generator parameters.
This dependency-ordered warm-up trains the generator against fixed
upstream representations.

%% file: tab/tables_synced.tex
\begin{table*}[t]
	\centering
	\caption{MSE of different online methods with horizons in \{24, 48, 96\}. ``$\backslash$'' denotes the offline-trained backbone without online updates; FSNet is built on TCN and is reported on that backbone only. The best results are marked in \textbf{bold} and the second best are \underline{underlined}.
 }\label{tab:main_result_mse}
  \resizebox{0.86\textwidth}{!}{
  \begin{tabular}{cc|ccc|ccc|ccc|ccc|ccc}
\toprule
 & \multirow{2}{*}{\diagbox[width=2cm,height=0.85cm,innerrightsep=2pt]{\textbf{\small Method}}{\small \textbf{Dataset}\\$H$}} & \multicolumn{3}{c|}{\textbf{ETTh2}} & \multicolumn{3}{c|}{\textbf{ETTm1}} & \multicolumn{3}{c|}{\textbf{Weather}} & \multicolumn{3}{c|}{\textbf{ECL}} & \multicolumn{3}{c}{\textbf{Traffic}} \\
\multicolumn{1}{c}{\textbf{Model}} &  & \textbf{24} & \textbf{48} & \textbf{96} & \textbf{24} & \textbf{48} & \textbf{96} & \textbf{24} & \textbf{48} & \textbf{96} & \textbf{24} & \textbf{48} & \textbf{96} & \textbf{24} & \textbf{48} & \textbf{96} \\ \midrule

\multirow{8}{*}{TCN}
 & \textbackslash{} & 3.156 & 4.141 & 6.019 & 1.770 & 1.908 & 1.673 & 3.301 & 9.718 & 8.588 & 57.782 & 57.948 & 58.729 & 0.569 & 0.637 & 0.676 \\
 & GD & 3.223 & 4.408 & 6.291 & 1.015 & 1.172 & 1.142 & 0.868 & 1.346 & 1.799 & 6.315 & 8.176 & 11.385 & 0.452 & 0.509 & 0.576 \\
 & FSNet & 3.079 & 4.247 & 6.213 & 0.763 & 0.923 & 1.003 & 0.875 & 1.338 & 1.755 & 6.143 & 9.435 & 12.941 & 0.458 & 0.517 & 0.574 \\
 & OneNet & \underline{2.965} & 4.892 & 8.257 & 1.335 & 1.547 & 1.040 & 0.861 & \textbf{1.182} & \textbf{1.484} & 9.757 & 10.847 & 15.932 & 0.462 & 0.544 & 0.591 \\
 & SOLID++ & 3.231 & 4.111 & 6.191 & 0.674 & 0.795 & 0.868 & 0.867 & 1.311 & 1.670 & 5.991 & \underline{7.340} & \textbf{9.412} & \textbf{0.418} & 0.479 & \underline{0.529} \\
 & DSOF & 3.074 & 5.659 & 8.670 & 0.616 & 0.811 & 0.982 & 0.953 & 1.294 & 1.662 & \textbf{5.308} & 8.050 & 10.860 & 0.420 & \underline{0.460} & 0.574 \\
 & PROCEED & \underline{2.965} & \underline{4.093} & \underline{5.952} & \underline{0.524} & \underline{0.691} & \underline{0.792} & \underline{0.816} & 1.233 & 1.703 & 6.025 & 7.510 & 10.470 & 0.420 & \underline{0.460} & 0.574 \\
 & \modelns & \textbf{2.876} & \textbf{4.023} & \textbf{5.908} & \textbf{0.512} & \textbf{0.689} & \textbf{0.791} & \textbf{0.810} & \underline{1.205} & \underline{1.583} & \underline{5.310} & \textbf{7.334} & \underline{10.090} & \underline{0.419} & \textbf{0.456} & \textbf{0.505} \\ \midrule

\multirow{7}{*}{PatchTST}
 & \textbackslash{} & 2.880 & 4.103 & 6.178 & 0.549 & 0.726 & 0.769 & 0.746 & 1.006 & 1.311 & 4.207 & 4.942 & 6.042 & 0.378 & 0.386 & 0.403 \\
 & GD & 2.092 & \underline{3.434} & \textbf{5.770} & 0.455 & 0.598 & 0.674 & 0.735 & \underline{0.979} & \underline{1.261} & 4.143 & 4.762 & 5.791 & 0.376 & 0.380 & 0.398 \\
 & OneNet & 2.717 & 4.095 & 6.044 & 0.681 & 0.721 & 0.788 & 0.824 & 1.052 & 1.319 & 4.111 & 4.752 & 5.767 & 0.360 & 0.372 & 0.387 \\
 & SOLID++ & 2.648 & 3.925 & 6.148 & 0.447 & \underline{0.580} & \textbf{0.659} & 0.735 & 0.981 & 1.262 & 4.156 & 4.780 & 5.835 & 0.376 & 0.378 & 0.397 \\
 & DSOF & 9.761 & 10.840 & 6.150 & 0.684 & 0.847 & 0.974 & 1.199 & 1.598 & 1.975 & 4.582 & \underline{4.584} & 10.110 & 0.475 & 0.426 & 0.534 \\
 & PROCEED & \underline{1.977} & 3.462 & 6.828 & \underline{0.429} & 0.587 & 0.705 & \underline{0.729} & 0.981 & 1.262 & \underline{3.935} & 4.718 & \underline{5.656} & \underline{0.339} & \underline{0.358} & \underline{0.376} \\
 & \modelns & \textbf{1.858} & \textbf{3.349} & \underline{6.032} & \textbf{0.423} & \textbf{0.561} & \underline{0.671} & \textbf{0.727} & \textbf{0.965} & \textbf{1.260} & \textbf{3.917} & \textbf{4.481} & \textbf{5.642} & \textbf{0.334} & \textbf{0.356} & \textbf{0.356} \\ \midrule

\multirow{7}{*}{iTransformer}
 & \textbackslash{} & 3.605 & 4.992 & 7.114 & 0.603 & 0.807 & 0.832 & 0.942 & 1.211 & 1.461 & 4.082 & 4.874 & 6.054 & 0.354 & 0.372 & 0.393 \\
 & GD & 2.637 & 4.148 & 6.734 & 0.468 & 0.606 & 0.695 & 0.869 & 1.143 & 1.409 & 4.055 & 4.851 & 5.989 & 0.348 & 0.369 & 0.388 \\
 & OneNet & 2.985 & \textbf{4.015} & 6.451 & 0.609 & 0.746 & 0.766 & 0.803 & 1.089 & 1.418 & 4.077 & 4.906 & 6.100 & 0.349 & 0.371 & 0.395 \\
 & SOLID++ & 2.804 & 4.278 & 6.582 & 0.455 & 0.601 & 0.688 & 0.875 & 1.139 & 1.403 & 4.070 & 4.816 & 6.024 & 0.342 & 0.365 & 0.384 \\
 & DSOF & 2.875 & 4.356 & 6.548 & 0.649 & 0.876 & 1.045 & 1.019 & 1.431 & 1.809 & 5.072 & 6.338 & 7.982 & 0.420 & 0.457 & 0.478 \\
 & PROCEED & \underline{2.472} & 4.156 & \underline{6.233} & \underline{0.425} & \underline{0.559} & \underline{0.646} & \underline{0.753} & \underline{1.011} & \underline{1.289} & \textbf{3.863} & \underline{4.631} & \underline{5.816} & \underline{0.329} & \underline{0.352} & \underline{0.373} \\
 & \modelns & \textbf{2.227} & \underline{4.080} & \textbf{6.232} & \textbf{0.421} & \textbf{0.556} & \textbf{0.643} & \textbf{0.733} & \textbf{1.004} & \textbf{1.277} & \underline{3.872} & \textbf{4.627} & \textbf{5.573} & \textbf{0.327} & \textbf{0.343} & \textbf{0.371} \\ \bottomrule
\end{tabular}
}
\end{table*}

\begin{table*}[t]
	\centering
	\caption{MAE of different online methods with horizons in \{24, 48, 96\}; notation as in Table~\ref{tab:main_result_mse}.
 }\label{tab:main_result_mae}
  \resizebox{0.86\textwidth}{!}{
  \begin{tabular}{cc|ccc|ccc|ccc|ccc|ccc}
\toprule
 & \multirow{2}{*}{\diagbox[width=2cm,height=0.85cm,innerrightsep=2pt,innerwidth=1.5cm]{\textbf{\small Method}}{\small \textbf{Dataset}\\$H$}} & \multicolumn{3}{c|}{\textbf{ETTh2}} & \multicolumn{3}{c|}{\textbf{ETTm1}} & \multicolumn{3}{c|}{\textbf{Weather}} & \multicolumn{3}{c|}{\textbf{ECL}} & \multicolumn{3}{c}{\textbf{Traffic}} \\
\multicolumn{1}{c}{\textbf{Model}} &  & \textbf{24} & \textbf{48} & \textbf{96} & \textbf{24} & \textbf{48} & \textbf{96} & \textbf{24} & \textbf{48} & \textbf{96} & \textbf{24} & \textbf{48} & \textbf{96} & \textbf{24} & \textbf{48} & \textbf{96} \\ \midrule

\multirow{8}{*}{TCN}
 & \textbackslash{} & 0.729 & 0.800 & 0.921 & 0.771 & 0.819 & 0.782 & 0.819 & 1.909 & 1.453 & 0.659 & 0.676 & 0.693 & 0.351 & 0.392 & 0.404 \\
 & GD & 0.727 & 0.824 & 0.946 & 0.635 & 0.674 & 0.667 & 0.406 & 0.561 & 0.703 & 0.401 & 0.446 & 0.488 & 0.320 & 0.350 & 0.383 \\
 & FSNet & 0.705 & 0.829 & 0.957 & 0.521 & 0.581 & 0.609 & 0.406 & 0.558 & 0.690 & 0.390 & 0.436 & 0.474 & 0.325 & 0.355 & 0.384 \\
 & OneNet & 0.728 & 0.902 & 1.096 & 0.688 & 0.781 & 0.678 & 0.456 & 0.591 & 0.707 & 0.767 & 0.721 & 0.505 & 0.319 & 0.366 & 0.393 \\
 & SOLID++ & 0.711 & 0.799 & 0.916 & 0.524 & 0.566 & 0.590 & 0.398 & 0.543 & \textbf{0.667} & 0.394 & 0.430 & 0.472 & 0.308 & 0.334 & 0.353 \\
 & DSOF & 0.812 & 1.018 & 1.230 & 0.476 & 0.550 & 0.623 & 0.473 & 0.602 & 0.715 & 0.405 & 0.438 & 0.474 & \underline{0.298} & \underline{0.314} & 0.344 \\
 & PROCEED & \underline{0.665} & \underline{0.761} & \underline{0.897} & \underline{0.447} & \underline{0.520} & \underline{0.553} & \underline{0.385} & \underline{0.527} & 0.674 & \underline{0.388} & \underline{0.428} & \underline{0.468} & \underline{0.298} & 0.315 & \underline{0.326} \\
 & \modelns & \textbf{0.656} & \textbf{0.754} & \textbf{0.896} & \textbf{0.445} & \textbf{0.513} & \textbf{0.548} & \textbf{0.382} & \textbf{0.523} & \underline{0.672} & \textbf{0.383} & \textbf{0.419} & \textbf{0.462} & \textbf{0.276} & \textbf{0.313} & \textbf{0.321} \\ \midrule

\multirow{7}{*}{PatchTST}
 & \textbackslash{} & 0.684 & 0.770 & 0.891 & 0.450 & 0.528 & 0.550 & 0.373 & 0.491 & 0.612 & 0.297 & 0.321 & 0.346 & 0.277 & 0.276 & 0.286 \\
 & GD & \underline{0.611} & \underline{0.718} & 0.864 & 0.407 & 0.474 & 0.509 & 0.369 & \underline{0.482} & \underline{0.592} & 0.294 & 0.315 & 0.341 & 0.276 & 0.267 & 0.278 \\
 & OneNet & 0.631 & 0.731 & \underline{0.862} & 0.477 & 0.504 & 0.541 & 0.392 & 0.496 & 0.605 & \underline{0.291} & \underline{0.314} & \underline{0.340} & 0.259 & 0.259 & 0.263 \\
 & SOLID++ & 0.670 & 0.761 & 0.890 & 0.406 & \underline{0.467} & \underline{0.504} & 0.372 & 0.485 & 0.597 & 0.294 & 0.315 & \underline{0.340} & 0.276 & 0.264 & 0.277 \\
 & DSOF & 1.115 & 1.206 & 1.604 & 0.545 & 0.603 & 0.650 & 0.634 & 0.777 & 0.884 & 0.570 & 0.613 & 0.601 & 0.345 & 0.373 & 0.375 \\
 & PROCEED & \underline{0.611} & 0.733 & 0.926 & \underline{0.395} & 0.468 & 0.517 & \underline{0.367} & \underline{0.482} & 0.593 & 0.294 & 0.318 & 0.341 & \underline{0.235} & \underline{0.244} & \underline{0.249} \\
 & \modelns & \textbf{0.589} & \textbf{0.701} & \textbf{0.858} & \textbf{0.392} & \textbf{0.454} & \textbf{0.502} & \textbf{0.364} & \textbf{0.480} & \textbf{0.591} & \textbf{0.288} & \textbf{0.313} & \textbf{0.336} & \textbf{0.230} & \textbf{0.242} & \textbf{0.241} \\ \midrule

\multirow{7}{*}{iTransformer}
 & \textbackslash{} & 0.724 & 0.822 & 0.944 & 0.478 & 0.566 & 0.584 & 0.473 & 0.581 & 0.676 & 0.289 & 0.315 & 0.344 & 0.257 & 0.267 & 0.280 \\
 & GD & 0.669 & 0.788 & 0.943 & 0.422 & 0.488 & 0.529 & 0.444 & 0.552 & 0.651 & 0.287 & 0.313 & 0.341 & 0.251 & 0.262 & 0.272 \\
 & OneNet & 0.663 & 0.775 & 0.916 & 0.449 & 0.525 & 0.546 & 0.411 & 0.526 & 0.633 & 0.289 & 0.316 & 0.345 & 0.252 & 0.264 & 0.277 \\
 & SOLID++ & 0.673 & 0.789 & 0.929 & 0.421 & 0.488 & 0.527 & 0.447 & 0.554 & 0.653 & 0.288 & 0.314 & 0.343 & 0.248 & 0.260 & 0.269 \\
 & DSOF & 0.647 & \underline{0.749} & 0.902 & 0.471 & 0.559 & 0.627 & 0.482 & 0.631 & 0.745 & 0.358 & 0.412 & 0.456 & 0.279 & 0.294 & 0.301 \\
 & PROCEED & \underline{0.637} & 0.774 & \underline{0.892} & \underline{0.397} & \underline{0.467} & \textbf{0.502} & \underline{0.385} & \underline{0.496} & \underline{0.604} & \underline{0.280} & \underline{0.306} & \underline{0.334} & \underline{0.229} & \textbf{0.242} & \underline{0.254} \\
 & \modelns & \textbf{0.626} & \textbf{0.742} & \textbf{0.874} & \textbf{0.382} & \textbf{0.461} & \underline{0.516} & \textbf{0.375} & \textbf{0.492} & \textbf{0.603} & \textbf{0.278} & \textbf{0.302} & \textbf{0.332} & \textbf{0.221} & \underline{0.243} & \textbf{0.252} \\ \bottomrule
\end{tabular}
}
\end{table*}

%% file: sec/5.experiment.tex
\section{Experiments}
We evaluate \textbf{Unmix Drift} under delayed feedback: we first describe
the baselines, datasets, and leakage-free protocol, then present the main
results, ablations, mechanism analysis, and online cost.

\subsection{Baselines and Datasets}
\label{sec:baselines_datasets}

\textbf{Baselines.}
We compare with online gradient descent (GD) and five online adaptation
methods: FSNet~\cite{pham2023learning}, a fast--slow framework that adjusts
convolutional filters of a TCN and is therefore reported on that backbone;
OneNet~\cite{wen2023onenet}, which online-ensembles multiple forecasters;
SOLID++~\cite{chen2024calibration}, the continually fine-tuned variant of
SOLID with context-driven calibration; DSOF~\cite{lau2025fast}, a dual-stream
method that uses pseudo labels for not-yet-observed targets; and
PROCEED~\cite{zhao2025proactive}, which proactively adapts from a global
drift representation without label feedback. Baseline implementations and
settings follow PROCEED.

\textbf{Datasets.}
We use five real-world multivariate datasets: ETTh2 and
ETTm1~\cite{zhou2021informer} (electricity transformer temperature at hourly
and 15-minute resolution), Weather~\cite{zhou2021informer} (meteorological
variables with seasonal patterns and gradual shifts),
ECL~\cite{zeng2023transformers} (hourly consumption of hundreds of clients
with strong cross-variable dependencies), and
Traffic~\cite{wu2020connecting} (freeway occupancy rates; large-scale and
highly non-stationary). Following~\cite{zhao2025proactive}, each dataset is
split chronologically into training, validation, and test sets, and online
learning runs over the validation and test streams.

\subsection{Experimental Settings}
\label{sec:exp_settings}

\textbf{Leakage-free delayed-feedback protocol.}
At each online step $t$, the newly observed value completes the delayed
target $\mathbf{Y}_{t-H}$; the model first performs the causal base update of
Eq.~\eqref{eq:causal_base_update} with $(\mathbf{X}_{t-H},\mathbf{Y}_{t-H})$,
constructs $\mathbf{D}_t^-$, and then forecasts $\widehat{\mathbf{Y}}_t$ from the
still-unlabeled query $\mathbf{X}_t$. The query-conditioned modulation is
temporary and uses no element of $\mathbf{Y}_t$, which remains unavailable
until $t+H$. This protocol applies to \textbf{all} methods.

\textbf{Backbones and fairness.}
We evaluate on three backbones: TCN~\cite{bai2018empirical},
PatchTST~\cite{nie2022time}, and
iTransformer~\cite{liu2024itransformerinvertedtransformerseffective}. All
methods share the same backbone, data splits, delayed-feedback protocol, and,
unless otherwise stated, the same online update budget.

\textbf{Training, selection, and evaluation.}
The backbone and adaptation modules are initialized on the training split and
refined on the validation stream as described in
Section~\ref{sec:stage_warmup}. After model selection on the validation stream,
all hyperparameters and warm-up stage boundaries are frozen. The resulting
model state is then carried into a single chronological pass over the test
stream. During this pass, test labels are incorporated into causal online
updates only when their corresponding forecast targets become available
under the prescribed delay. All reported metrics are computed from
forecasts issued during this test pass.

\textbf{Optimization and metrics.}
All methods are implemented in PyTorch with AdamW. Learning rates, weight
decay, and online update learning rates are tuned on the validation stream
per dataset--backbone pair, and early stopping is applied to offline training
where applicable. The architectural configuration, normalization pipeline,
delayed-update count, temperature $\tau$, and the number of components
($K{=}2$ throughout; see Section~\ref{sec:ablation}) are held fixed across
the test stream. We report MSE and MAE over the prediction horizon, computed
at every online step and aggregated over the whole stream, averaged over
3 random seeds.

\input{tab/ablation_merged}

\begin{figure}[t]
    \centering
    \includegraphics[width=0.80\columnwidth]{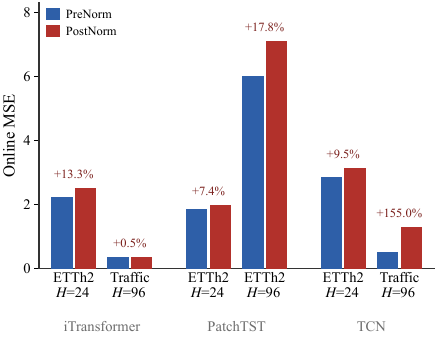}
    \Description{A grouped bar chart comparing PreNorm and PostNorm
    salience estimation across six forecasting configurations.
    PostNorm has higher MSE in all six configurations, with
    relative increases ranging from 0.5 to 155 percent.}
    \caption{
        PreNorm vs.\ PostNorm salience estimation.
        Bars show three-seed mean MSE; percentages denote the
        relative MSE increase of PostNorm over PreNorm.
    }
    \label{fig:postnorm}
\end{figure}

\subsection{Main Results}
\label{sec:main_results}

Table~\ref{tab:main_result_mse} and Table~\ref{tab:main_result_mae} report the
main results under the leakage-free delayed-feedback protocol. Unmix Drift
ranks first or second in all 45 backbone--dataset--horizon settings on both
MSE (first in 36) and MAE (first in 42), and improves over the strongest
proactive baseline PROCEED by 2.7\% MSE on average and by up to 12.0\%
(TCN, Traffic, $H{=}96$). Here and below, relative improvements are
macro-averaged over the corresponding backbone--dataset--horizon
configurations.

\begin{figure*}[t]
    \centering
    \includegraphics[width=0.76\textwidth]{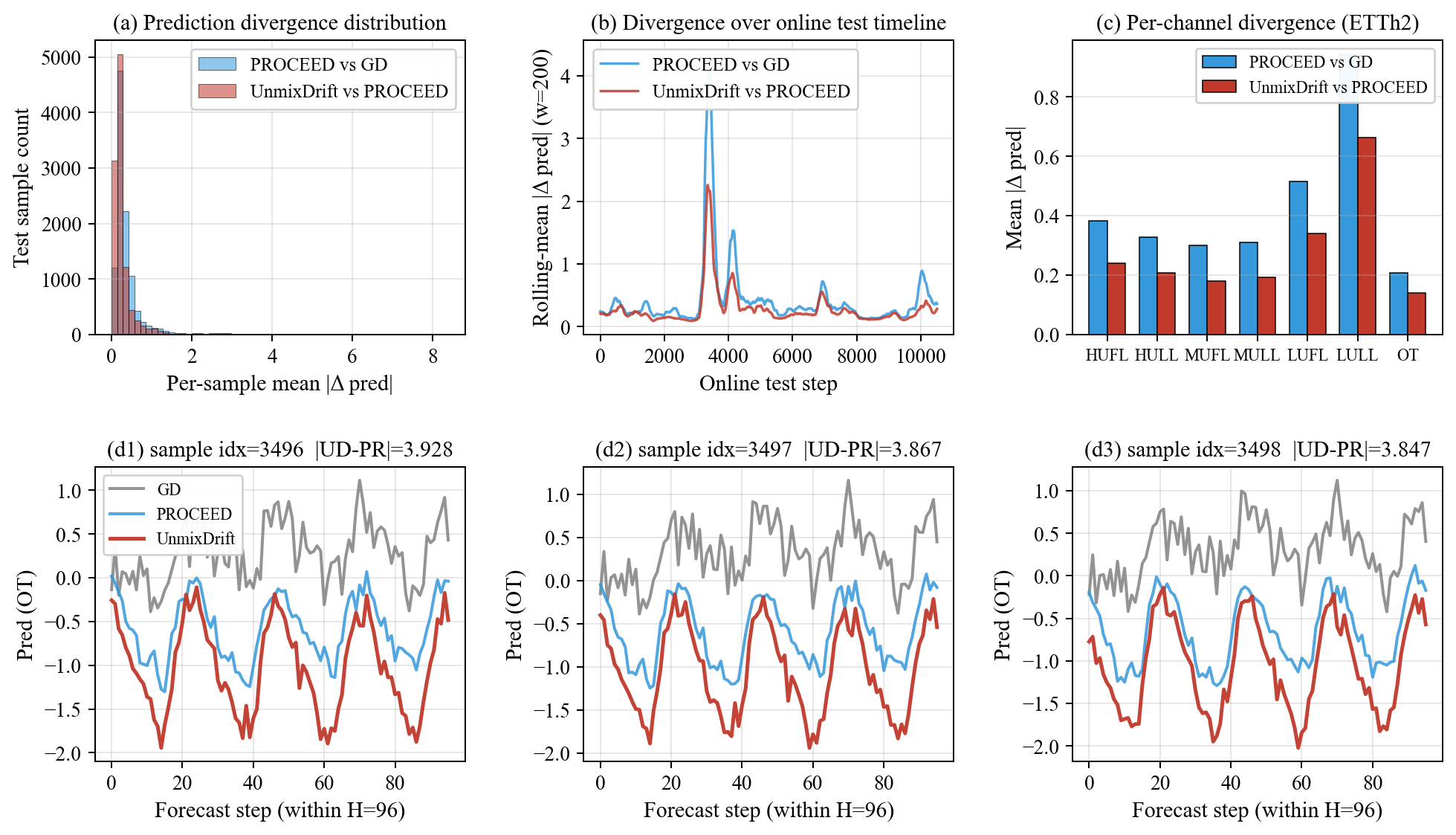}
    \Description{
        Six-panel prediction-divergence analysis on ETTh2.
        Panels (a)--(c) compare prediction differences
        between PROCEED and GD, and between Unmix Drift
        and PROCEED, across samples, time, and channels.
        Panels (d1)--(d3) show forecasts for three
        consecutive high-response samples.
    }
    \caption{
        Prediction-divergence diagnostics on ETTh2
        ($H{=}96$, iTransformer).
        (a)~Distribution of per-sample prediction divergence between
        method pairs.
        (b)~Rolling-mean divergence over the online test stream
        (window 200).
        (c)~Per-channel mean divergence.
        (d1--d3)~Forecasts for three consecutive high-response samples;
        $|\mathrm{UD}-\mathrm{PR}|$ denotes the per-sample divergence
        between Unmix Drift and PROCEED.
    }
    \label{fig:pred_divergence}
\end{figure*}

\textbf{(O1): Proactive adaptation is consistently competitive under delayed
feedback.}
In 37 of 45 settings, the best MSE is achieved by a proactive method
(Unmix Drift or PROCEED); methods relying on delayed labels or
prediction-based surrogates (GD, FSNet, OneNet, SOLID++, and DSOF) lead in
the remaining 8, where Unmix Drift still ranks second. These results are
consistent with the motivation for proactive adaptation: when complete
supervision is delayed, the current unlabeled query provides an additional
adaptation signal before its corresponding target becomes available.
Accordingly, proactive adaptation remains highly competitive across
backbones, datasets, and forecasting horizons under the same causal protocol.

\textbf{(O2): Component-wise drift construction consistently improves over
the global counterpart.}
PROCEED and Unmix Drift operate under the same backbones, delayed-feedback
protocol, online update budget, and proactive generate-and-modulate
paradigm, while Unmix Drift replaces the globally aggregated discrepancy
with the component-wise construction in Section~\ref{subsec:prototype_unmixing}.
Across this comparison, Unmix Drift attains lower MSE in 44 of 45 settings,
with the only exception (iTransformer, ECL, $H{=}24$) differing by less than
0.3\%. The largest margins are 12.0\% (TCN, Traffic, $H{=}96$), 11.9\%
(TCN, ECL, $H{=}24$), and 11.7\% (PatchTST, ETTh2, $H{=}96$).
This consistent pattern supports preserving the discrepancy structure before
final aggregation rather than forming a single global difference at the
outset. For the $K{=}2$ model used throughout the main experiments, the
construction can be interpreted more specifically as a signed redistribution
direction modulated by component salience: normalized residuals determine
the direction of the current--resolved-context redistribution, whereas
pre-normalization relative changes determine how strongly different
components contribute to the resulting drift signal.

\textbf{(O3): The gains generalize across backbone families and vary across
datasets.}
The average improvement over PROCEED is observed for all three backbone
families---3.3\% on TCN, 3.1\% on PatchTST, and 1.7\% on iTransformer---so
the benefit is not restricted to a particular forecasting architecture.
Across datasets, the corresponding gains are 4.2\% on ETTh2, 3.1\% on ECL,
2.7\% on Traffic, 1.8\% on Weather, and 1.7\% on ETTm1. Thus, the magnitude
of the improvement is dataset dependent, with larger gains on ETTh2, ECL,
and Traffic in the present evaluation. This variation suggests that the
benefit of retaining component-wise discrepancy structure may depend on the
form and magnitude of non-stationarity in the underlying stream.

\subsection{Ablation Study}
\label{sec:ablation}

We examine five design choices: drift aggregation,
salience estimation, global gating and capacity,
stage-wise warm-up, and component count $K$.
Unless stated otherwise, each variant changes only
the designated part of the framework.
Table~\ref{tab:ablation} reports MSE averaged over
$H\in\{24,48,96\}$ on iTransformer and PatchTST;
its \emph{Full} column matches
Table~\ref{tab:main_result_mse}.

\textbf{How should component-wise residuals be aggregated?}
\emph{Mean} replaces the salience weights in
Eq.~\eqref{eq:prototype_drift_weight} with uniform
averaging, while \emph{Max} retains only the
strongest component. Full achieves the lowest MSE
across all 10 backbone--dataset pairs. Mean
increases average MSE by 2.5\% on iTransformer
and 8.1\% on PatchTST, while Max causes larger
degradation, particularly on PatchTST/ECL.
These results favor salience-weighted aggregation
over uniform averaging or hard selection.

\textbf{Should salience be estimated before normalization?}
PostNorm replaces pre-normalization responses with
normalized mixing scores for salience estimation,
leaving the remaining pipeline unchanged.
It yields higher MSE in all six configurations,
with the same 6/6 pattern on MAE. Excluding the
largest degradation (TCN/Traffic, $H{=}96$),
the average MSE increase remains 9.7\%.
These results support pre-normalization
salience estimation.

\input{tab/global}

\textbf{Can global gating and matched capacity explain the gain?}
We construct a capacity-matched \emph{GatedGlobal}
control using a global current--reference
discrepancy, learned gating, and the same adaptation
generator as Unmix Drift. Across the six configurations,
with PROCEED and Unmix Drift values from the main
experiments, Unmix Drift achieves lower MSE and MAE
than GatedGlobal, reducing MSE by 7.1\% on average.
GatedGlobal shows mixed results relative to PROCEED.
Thus, gating and matched adaptation capacity alone
do not explain the gains from component-wise
drift construction.

\textbf{Does the training schedule matter?}
\emph{Joint} optimizes all modules using
$\mathcal{L}_{\mathrm{forecast}}+
\mathcal{L}_{\mathrm{unmix}}$.
\emph{Skip Stage~1} removes projection warm-up,
while \emph{Skip Stage~2} also removes the
reconstruction refinement in
Eq.~\eqref{eq:unmix_loss}.
All variants increase MSE across the 10
backbone--dataset pairs. Joint increases average
MSE by 2.8\% on iTransformer and 7.0\% on
PatchTST; removing Stage~1 or Stage~2 increases
PatchTST MSE by 11.2\% and 9.0\%, respectively.
Since Skip Stage~2 also changes its objective,
it is a combined ablation. These results support
dependency-ordered warm-up.

\textit{Number of components.}
We evaluate $K\in\{1,2,4,8,16\}$ at $H{=}24$
on iTransformer/ETTm1, PatchTST/ETTm1, and
TCN/ETTh2 (3 seeds each).
$K{=}2$ achieves the lowest mean MSE in all
three settings. Both $K{=}1$ and larger $K$
perform worse on average. We therefore use
$K{=}2$ as a parsimonious non-trivial
decomposition without assuming it is universally optimal.

\begin{figure}[t]
    \centering
    \includegraphics[width=0.85\linewidth]{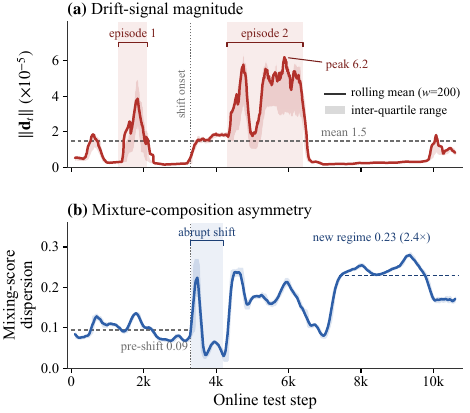}
    \Description{
    Two-panel diagnostics of the K=2 model on
    iTransformer/ETTh2 at H=96. Panel (a) shows
    drift-signal magnitude over online test steps,
    including two elevated episodes and a peak near
    6.2 times 10^{-5}. Panel (b) shows mixing-score
    dispersion across components, increasing from
    approximately 0.09 before the shift to 0.23
    in the new regime.
    }    
    \caption{
    Diagnostics of the $K=2$ model on iTransformer/ETTh2
    ($H=96$; rolling window 200).
    (a) Drift-signal magnitude with its interquartile range.
    (b) Mixing-score dispersion across components,
    showing a change from approximately 0.09 to 0.23.
    }
    \label{fig:proto_dynamics}
\end{figure}

\textbf{Prediction-level diagnostics.}
Figure~\ref{fig:pred_divergence} shows that the
Unmix Drift--PROCEED divergence is generally smaller
than the PROCEED--GD divergence in sample-wise
and per-channel comparisons. Localized peaks
appear along the test stream, while consecutive
high-response samples exhibit coherent changes
across the forecast horizon. These observations
suggest selective refinement of proactive
forecasts during particular online episodes.

\textbf{Adaptation-signal diagnostics.}
Figure~\ref{fig:proto_dynamics} examines the
learned $K{=}2$ signal on iTransformer/ETTh2
at $H{=}96$. Drift magnitude increases during
localized episodes, indicating stronger
adaptation signals at these points. Meanwhile,
mixing-score dispersion across components rises
near the stronger episode, increasing from
approximately $0.09$ to $0.23$, and later
stabilizes at a higher level. Together, these
patterns show that the learned signal varies
both in overall magnitude and in how its
contributions are distributed across components.

\subsection{Efficiency Analysis}

Table~\ref{tab:efficiency} reports batch-size-1 per-step latency
(trimmed mean after 50 warm-up steps), peak PyTorch GPU memory,
and additional trainable parameters for three representative
backbone--dataset pairs at $H=24$. Forecasting performance
follows the three-seed results in Table~1.

Relative to GD, both proactive methods incur 1.6--2.6$\times$
total latency and 1.2--1.6$\times$ peak memory. Compared with
PROCEED, Unmix Drift changes total latency by $-1.4\%$ to
$+6.7\%$ and peak memory by $+0.5\%$ to $+4.7\%$, despite
using 0.8--3.5M additional trainable parameters versus
0.4--0.7M. The inference and update breakdown shows that
these differences remain modest across the evaluated settings.
Notably, on iTransformer--Traffic, Unmix Drift uses 3.53M
adaptation parameters versus 0.64M for PROCEED, yet reduces
total latency from 53.737 ms to 53.001 ms, with a 30.8 MB
increase in peak memory. Overall, the additional parameters
incur only modest measured overhead in the evaluated settings.

\input{tab/Efficiency_table}

%% file: tab/ablation_merged.tex


\definecolor{UMBlue}{HTML}{3F6682}
\definecolor{UMBlueLight}{HTML}{EEF4F8}
\definecolor{UMGroup}{HTML}{F3F6F8}
\definecolor{DeltaRed}{HTML}{A85C5C}

\begin{table}[t]
    \centering

    \caption{
        Ablation of drift aggregation and stage-wise warm-up.
        MSE is averaged over horizons $\{24,48,96\}$;
        $\Delta$ denotes the average relative increase over the full model.
    }

    \label{tab:ablation}

    \footnotesize
    \renewcommand{\arraystretch}{1.08}

    \begin{tabular*}{\columnwidth}{
        @{\extracolsep{\fill}}
        c
        >{\columncolor{UMBlueLight}}c
        c c
        c c c
        @{}
    }


        \toprule

        \multirow{2}{*}{\textbf{Dataset}}
        &
        \multirow{2}{*}{\textbf{Full}}
        &
        \multicolumn{2}{c}{
            \textcolor{UMBlue}{\textbf{Aggregation}}
        }
        &
        \multicolumn{3}{c}{
            \textcolor{UMBlue}{\textbf{Warm-up}}
        }
        \\

        \cmidrule(lr){3-4}
        \cmidrule(l){5-7}

        &
        &
        Mean
        &
        Max
        &
        Joint
        &
        w/o S1
        &
        w/o S2
        \\

        \midrule


        \rowcolor{UMGroup}
        \multicolumn{7}{@{}c@{}}{
            \textcolor{UMBlue}{\textbf{iTransformer}}
        }
        \\[-1pt]

        ETTh2
        &
        \textbf{4.180}
        &
        4.342
        &
        4.944
        &
        4.306
        &
        4.376
        &
        4.359
        \\

        ETTm1
        &
        \textbf{0.540}
        &
        0.554
        &
        0.562
        &
        0.554
        &
        0.551
        &
        0.555
        \\

        Weather
        &
        \textbf{1.005}
        &
        1.023
        &
        1.038
        &
        1.013
        &
        1.020
        &
        1.020
        \\

        ECL
        &
        \textbf{4.691}
        &
        4.806
        &
        4.986
        &
        4.988
        &
        4.811
        &
        5.006
        \\

        Traffic
        &
        \textbf{0.347}
        &
        0.353
        &
        0.353
        &
        0.352
        &
        0.352
        &
        0.353
        \\


        \addlinespace[1pt]
        \cmidrule(lr){1-7}

        \textit{Avg.~$\Delta$}
        &
        --
        &
        \textcolor{DeltaRed}{+2.5\%}
        &
        \textcolor{DeltaRed}{+6.7\%}
        &
        \textcolor{DeltaRed}{+2.8\%}
        &
        \textcolor{DeltaRed}{+2.4\%}
        &
        \textcolor{DeltaRed}{+3.4\%}
        \\


        \midrule


        \rowcolor{UMGroup}
        \multicolumn{7}{@{}c@{}}{
            \textcolor{UMBlue}{\textbf{PatchTST}}
        }
        \\[-1pt]

        ETTh2
        &
        \textbf{3.746}
        &
        4.466
        &
        4.438
        &
        4.175
        &
        4.480
        &
        4.466
        \\

        ETTm1
        &
        \textbf{0.552}
        &
        0.589
        &
        0.596
        &
        0.592
        &
        0.594
        &
        0.595
        \\

        Weather
        &
        \textbf{0.984}
        &
        1.068
        &
        1.149
        &
        1.047
        &
        1.069
        &
        1.061
        \\

        ECL
        &
        \textbf{4.680}
        &
        4.863
        &
        17.960
        &
        5.049
        &
        5.511
        &
        5.066
        \\

        Traffic
        &
        \textbf{0.349}
        &
        0.356
        &
        0.357
        &
        0.356
        &
        0.357
        &
        0.356
        \\


        \addlinespace[1pt]
        \cmidrule(lr){1-7}

        \textit{Avg.~$\Delta$}
        &
        --
        &
        \textcolor{DeltaRed}{+8.1\%}
        &
        \textcolor{DeltaRed}{+65.9\%}
        &
        \textcolor{DeltaRed}{+7.0\%}
        &
        \textcolor{DeltaRed}{+11.2\%}
        &
        \textcolor{DeltaRed}{+9.0\%}
        \\

        \bottomrule

    \end{tabular*}

\end{table}

%% file: tab/global.tex

\begin{table}[t]
    \centering

    \caption{
        Capacity-matched global control.
        GatedGlobal augments a global current--reference discrepancy with
        learned gating while matching the adaptation capacity of Unmix Drift.
        Results are three-seed mean MSE; Gain denotes the relative reduction
        of Unmix Drift over GatedGlobal.
    }

    \label{tab:gated_global}

    \footnotesize
    \renewcommand{\arraystretch}{1.08}
    \setlength{\tabcolsep}{3.5pt}

    \begin{tabular*}{\columnwidth}{
        @{\extracolsep{\fill}}
        l
        c
        c
        c
        >{\columncolor{UMBlueLight}}c
        r
        @{}
    }

        \toprule

        \multirow{2}{*}{\textbf{Backbone}}
        &
        \multirow{2}{*}{\textbf{Dataset / $H$}}
        &
        \multicolumn{2}{c}{\textcolor{UMBlue}{\textbf{Global representation}}}
        &
        \multirow{2}{*}{\textbf{Unmix}}
        &
        \multirow{2}{*}{\textbf{Gain}}
        \\

        \cmidrule(lr){3-4}

        &
        &
        PROCEED
        &
        GatedGlobal
        &
        &
        \\

        \midrule

        \multirow{2}{*}{iTransformer}
        &
        ETTh2 / 24
        &
        2.472
        &
        2.429
        &
        \textbf{2.227}
        &
        \textcolor{DeltaRed}{8.3\%}
        \\

        &
        Traffic / 96
        &
        0.373
        &
        0.373
        &
        \textbf{0.371}
        &
        \textcolor{DeltaRed}{0.5\%}
        \\

        \midrule

        \multirow{2}{*}{PatchTST}
        &
        ETTh2 / 24
        &
        1.977
        &
        2.050
        &
        \textbf{1.858}
        &
        \textcolor{DeltaRed}{9.4\%}
        \\

        &
        ETTh2 / 96
        &
        6.828
        &
        7.340
        &
        \textbf{6.032}
        &
        \textcolor{DeltaRed}{17.8\%}
        \\

        \midrule

        \multirow{2}{*}{TCN}
        &
        ETTh2 / 24
        &
        2.965
        &
        2.983
        &
        \textbf{2.876}
        &
        \textcolor{DeltaRed}{3.6\%}
        \\

        &
        Traffic / 96
        &
        0.574
        &
        0.522
        &
        \textbf{0.505}
        &
        \textcolor{DeltaRed}{3.3\%}
        \\

        \bottomrule

    \end{tabular*}

\end{table}

%% file: tab/Efficiency_table.tex

\begin{table}[t]
    \centering

    \caption{
        Per-step online cost at $H{=}24$ (batch size 1).
        We report latency, peak GPU memory, and additional
        trainable parameters. GD fully fine-tunes the backbone.
    }
    \label{tab:efficiency}

    \vspace{-3pt}

    \footnotesize
    \renewcommand{\arraystretch}{1.02}
    \setlength{\tabcolsep}{5pt}

    \begin{tabular}{
        @{}
        c
        c c c
        c
        c
        @{}
    }

        \toprule

        \multirow{2}{*}{\textbf{Method}}
        &
        \multicolumn{3}{c}{
            \textcolor{UMBlue}{\textbf{Latency (ms)}}
        }
        &
        \multirow{2}{*}{\textbf{Mem. (MB)}}
        &
        \multirow{2}{*}{\textbf{Params (M)}}
        \\

        \cmidrule(lr){2-4}

        &
        Inf.
        &
        Upd.
        &
        Total
        &
        &
        \\

        \midrule


        \rowcolor{UMGroup}[0pt][0pt]
        \multicolumn{6}{@{}c@{}}{
            \textcolor{UMBlue}{\textbf{iTransformer + ECL}}
        }
        \\[-1pt]

        GD
        & 3.616
        & 12.561
        & 16.177
        & 190.4
        & --
        \\

        PROCEED
        & 11.888
        & 28.705
        & 40.593
        & 293.8
        & 0.73
        \\

        \rowcolor{UMBlueLight}[0pt][0pt]
        \textbf{\modelns}
        & 12.511
        & 30.248
        & 42.759
        & 298.0
        & 1.14
        \\


        \midrule

        \rowcolor{UMGroup}[0pt][0pt]
        \multicolumn{6}{@{}c@{}}{
            \textcolor{UMBlue}{\textbf{iTransformer + Traffic}}
        }
        \\[-1pt]

        GD
        & 4.713
        & 16.357
        & 21.071
        & 526.6
        & --
        \\

        PROCEED
        & 15.357
        & 38.380
        & 53.737
        & 653.2
        & 0.64
        \\

        \rowcolor{UMBlueLight}[0pt][0pt]
        \textbf{\modelns}
        & 15.268
        & 37.733
        & 53.001
        & 684.0
        & 3.53
        \\


        \midrule

        \rowcolor{UMGroup}[0pt][0pt]
        \multicolumn{6}{@{}c@{}}{
            \textcolor{UMBlue}{\textbf{PatchTST + ECL}}
        }
        \\[-1pt]

        GD
        & 4.817
        & 13.813
        & 18.630
        & 982.6
        & --
        \\

        PROCEED
        & 8.822
        & 21.780
        & 30.602
        & 1238.9
        & 0.43
        \\

        \rowcolor{UMBlueLight}[0pt][0pt]
        \textbf{\modelns}
        & 9.696
        & 22.968
        & 32.664
        & 1244.7
        & 0.84
        \\

        \bottomrule

    \end{tabular}

    \vspace{-5pt}

\end{table}

%% file: sec/6.conclusion.tex
\section{Conclusion}
\label{sec:conclusion}

We present Unmix Drift for online time-series forecasting with
delayed feedback. It preserves component-wise current--historical
discrepancies, combines normalized mixing-score residuals with
pre-normalization salience, and applies dependency-ordered warm-up
to projection, decomposition, and generation. Across five datasets,
three backbones, and three horizons, it achieves lower MSE than
PROCEED in 44/45 configurations and ranks among the top two methods
in both MSE and MAE across all settings, with latency and GPU memory
comparable to PROCEED. Ablations support the effectiveness of the
drift construction, salience design, and stage-wise training,
highlighting the value of structured discrepancies for proactive
adaptation.